\documentclass[a4, journal]{IEEEtran}
\usepackage{epsfig,subfigure, float}
\usepackage{amssymb,amsmath,amsthm,graphics,wrapfig}
\makeindex

\begin{document}

\title{On Three Challenges of Artificial Living Systems and Embodied Evolution}
\author{Serge Kernbach\\
IPVS, University of Stuttgart, Germany, {\it serge.kernbach@ipvs.uni-stuttgart.de}
\vspace{-8mm}
}
\maketitle

\begin{abstract} Creating autonomous, self-supporting, self-replicating, sustainable systems is a great challenge. To some extent, understanding life means not only being able to create it from scratch, but also improving, supporting, saving it, or even making it even more advanced. This can be thought of as a long-term goal of living technologies and embodied evolution. Current research agenda targets several short- and middle-term steps towards achieving such a vision: connection of ICT and bio-/chemo- developments, advances in "soft" and "wet" robotics, integration of material science into developmental robotics, and potentially, addressing the self-replication in autonomous systems.\footnote{This position paper is prepared for the consultation workshop "Living technology/Artificial systems/Embodied Evolution", organized by European Commission, Brussels 10.11.2011}
\end{abstract}
\vspace{-3mm}

\section{Current Research}
\label{sec:currentRes}

Current research on the fields of Living Technologies, Artificial Systems and Embodied Evolution is highly multidisciplinary and is related to many different bio-chemical, micro-technological and robotic areas.

\textbf{(a)} Collective systems~\cite{KernbachHCR11} are one of the major research fields devoted to common principal and mechanisms underlying mechatronic and non-mechatronic embodiment of living systems: different collective effects, scalability, self-organization, self-assembling as well as development of common evaluation tests and benchmarks. Essential attention is paid to individual and collective cognition~\cite{Kornienko_S05a}, exploring bio-inspired principles, collective homeostatic regulation~\cite{Kornienko_S06b} as well as different applications in manufacturing~\cite{Kornienko_S03A}. It is also related to analytical approaches~\cite{Levi99}, for example collective decision making~\cite{Kornienko_OS01}, which target guided self-assembling and self-organization in real systems. Many works cover the notion of embodied evolution~\cite{Koenig07} and development of multi-cellular artificial structures, their self-development, adaptation, reliability and other aspects~\cite{Levi10}.

\textbf{(b)} Bio-hybrid technologies represent a large field of research, which also approaches artificial living systems. One of the methods here is the combination of cultured (living) neurons and robots \cite{Novellino07} to investigate the dynamical and adaptive properties of neural systems \cite{Reger00}. This work is also related to understanding of how information is encoded~\cite{Cozzi06} and processed within a living neural network \cite{DeMarse01}. The hybrid technology can be used for neuro-robotic interfaces, different applications of \emph{in vitro} neural networks~\cite{Miranda09} or for bidirectional interaction between the brain and the external environment in both collective and non-collective systems. Several research projects already addressed the problem of controlling autonomous robots by living neurons~\cite{MartinoiaNeuroBit04}. Promising research area in bio-hybrid systems is the synthetic biology and the integration of real bio-chemical and microbiological systems into technological developments; for example using bacterial cellular mechanisms as sensors, development of bacterial bio-hybrid materials~\cite{Martel:2009p12325}, molecular synthesis of polymers and biofuels, genome engineering, and more general fields and challenges of synthetic biology~\cite{Alterovitz09}.

\textbf{(c)} Chemo-hybrid systems are another field of research, which targets bottom-up approach towards artificial living systems. It is inspired by artificial chemistry~\cite{Dittrich01}, self-replicating systems~\cite{Rasmussen13022004}, using bio-chemical mechanisms for e.g. cognition~\cite{Dale10} as well as by a general field of material science. In several works, this approach is denoted as swarm chemistry~\cite{Sayama09}. Researchers hope that such systems will give answers to questions related to developmental models~\cite{Astor00}, chemical computation, self-assembly, self-replication, simple chemistry-based ecologies~\cite{Breyer97} or technological capabilities of creating large-scale functional patterns \cite{Yin08}.
\vspace{-2mm}

\section{Research Agenda and Challenges}

Currently, three possible scenarios for living technologies and embodied evolution can be distinguished. Firstly, further development of micro- and nano- mechatronics can make it possible to achieve an advanced functionality at bottom layers (such as  self-replication). This involves several different technologies from material science; we can denote this scenario as \emph{nano-mechatronic scenario}. Secondly, the complexity of upper levels (such as information processing) can be handled by bio-technologies through advancements in minimal cell projects and cellular programmability, so that we can expect an appearance of pure \emph{bio-synthetic systems}. Finally, both approaches can be merged so that to make use of their advantages. This approach is termed \emph{bio-hybrid scenario}, which combines  bio-chemical and mechatronic autonomous systems.

\textbf{Challenge I:} Three above mentioned scenarios mean a further development of micro- and nano- mechatronics, the growth of bio- and chemo- synthetic systems, the hybridization of robotics, and appearance of ``soft/wet" systems. Each of these developments has own challenges, promises and risks. However, independent of what the dominant future technology might be, we face a new problem of integration of methodologies, paradigms and approaches from different areas of biology, chemistry, material science and robotics. This new integration will require  re-structuring current research landscape, which will not only essentially change the way we think about synthetic systems, but also extend their scientific and technological boundaries. Material sciences, bottom up chemistry and genetic engineering are especially relevant for open-ended embodied evolution and unbounded self-development -- which are essential challenges for artificial living systems.

\textbf{Challenge II:} The \emph{hybrid scenario} can be considered as the most probable way for future autonomous systems. Here we can identify several open research questions, the most important from them: \emph{How the current ICT can be combined with bio-chemical developments?} This question is also known in other formulations, e.g. "\emph{programmability of bio-synthetic systems}", and is the key point in a series of other scientific and technological challenges, and to some extent, even in understanding principles of synthetic life. Many research initiatives addressed it; this represents a key aspect of the long-term research agenda.

\textbf{Challenge III:} Artificial living systems possess a potentially high degree of plasticity and can undergo a developmental drift. There are many reasons for this: long-term developmental independence and autonomous behavior, emergence of artificial sociality, mechanisms of evolutionary self-organization. Such systems are very flexible and adaptive, but they also massively increase their own degrees of freedom. New challenges in this area are related to a long-term controllability and predictability of "self-*", principles of making plastic purposeful systems, predictability of structural development and engineering of open-ended evolution. These challenges have a great impact on the human community in general (cf. the "Terminator" scenario) as well as in different areas of embodied evolution and living technologies.

From the viewpoint of a short-term and middle-term research agenda, it would make sense to undertake a step-wise transition from current mechatronic towards hybrid collective systems: below are some examples of open research questions:

\emph{- Which properties of materials are useful for collective systems?}

\emph{- Capabilities of a minimal cognition by using simple (even molecular) systems?}

\emph{- Self-replication: from "wet hardware" to "soft hardware"}

\emph{- Are there artificial structural elements that are "absolutely plastic" in the developmental sense, analogous to biological amino acids?}

\emph{- Is a "natural chemistry" (i.e., a high complexity of evolutionary processes) important for adaptability and self-development?}

\emph{- What are the driving forces of long-term developmental processes? Are they controllable? Is the embodied evolution controllable?}

\emph{- Is there any developmental drift due to emergence of artificial sociality and self-recognition?}

\emph{- Do artificial homeostasis and energetic survival lead to appearance of cognitive capabilities and to emergence of different self-phenomena (denoted as "self-*"): self-replication, self-development, self-recovering?}

This position paper points to the important requirement to hybrid technologies -- integration of different research fields -- which will be a vital challenge in coming years. This can be done only in large interdisciplinary teams, e.g. within large European projects. Creating such teams as well as a necessary equipment in addressing fundamentals of different synthetic systems belongs to essential points of the long-term research agenda.
\vspace{-4mm}

\tiny

%\bibliography{../bibl_sk,../own_bibl_sk}

\end{document}